\documentclass[10pt,twocolumn,letterpaper]{article}

\usepackage{wacv}
\makeatletter %
  \@namedef{ver@everyshi.sty}{}
\makeatother
\usepackage{times}
\usepackage{epsfig}

\def\wacvPaperID{1224} %
\wacvalgorithmstrack   %
\wacvfinalcopy %

\usepackage{mathtools}
\usepackage{graphicx}
\graphicspath{{img/}}
\usepackage[svgpath=img]{svg}
\usepackage{epstopdf}
\usepackage{pgf}
\usepackage{calc}
\newcommand\inputpgf[2]{{
\let\pgfimageWithoutPath\pgfimage
\renewcommand{\pgfimage}[2][]{\pgfimageWithoutPath[##1]{#1/##2}}
\input{#1/#2}
}}

\usepackage{todonotes}
\usepackage{soul}
\usepackage{etoolbox}
\usepackage{amsmath}

\usepackage{amssymb}
\usepackage{bm}
\usepackage[export]{adjustbox}
\usepackage{tabularx}
\newcolumntype{Y}{>{\centering\arraybackslash}X} 
\usepackage{diagbox}
\usepackage{tikz}
\usetikzlibrary{bayesnet}
\newlength{\state} 
\usepackage{url}
\usepackage{array}
\usepackage{textcomp}
\usepackage{xspace}
\usepackage{booktabs}
\usepackage{tablefootnote}
\usepackage{cite}
\usepackage{multirow}
\usepackage{makecell}

\usepackage{comment}
\usepackage{color}
\usepackage{xcolor}

\usepackage{amsthm}

\usepackage{epigraph, varwidth}

\renewcommand{\epigraphsize}{\small}
\renewcommand{\textflush}{flushright}
\renewcommand{\sourceflush}{flushright}
\newcommand{\epitextfont}{\itshape}
\newcommand{\episourcefont}{\scshape}

\makeatletter
\newsavebox{\epi@textbox}
\newsavebox{\epi@sourcebox}
\newlength\epi@finalwidth
\renewcommand{\epigraph}[2]{%
  \vspace{\beforeepigraphskip}
  {\epigraphsize\begin{\epigraphflush}
   \epi@finalwidth=\z@
   \sbox\epi@textbox{%
     \varwidth{\epigraphwidth}
     \begin{\textflush}\epitextfont#1\end{\textflush}
     \endvarwidth
   }%
   \epi@finalwidth=\wd\epi@textbox
   \sbox\epi@sourcebox{%
     \varwidth{\epigraphwidth}
     \begin{\sourceflush}\episourcefont#2\end{\sourceflush}%
     \endvarwidth
   }%
   \ifdim\wd\epi@sourcebox>\epi@finalwidth 
     \epi@finalwidth=\wd\epi@sourcebox
   \fi
   \leavevmode\vbox{
     \hb@xt@\epi@finalwidth{\hfil\box\epi@textbox}
     \vskip1.75ex
     \hrule height \epigraphrule
     \vskip.75ex
     \hb@xt@\epi@finalwidth{\hfil\box\epi@sourcebox}
   }%
   \end{\epigraphflush}
   \vspace{\afterepigraphskip}}}
\makeatother
 
\usepackage{makeidx}

\usepackage{hyperref}
\hypersetup{
    colorlinks,
    linkcolor={red!50!black},
    citecolor={blue!50!black},
    urlcolor={blue!80!black}
}
\usepackage[capitalize]{cleveref}
\usepackage{xspace}
\renewcommand{\Cref}[1]{\cref{#1}} 
\usepackage{tabularx} 

\newcommand{\optional}[1]{}
\newcommand{\veryOptional}[1]{}
\newcommand{\hideout}[1]{}
\newcommand{\blue}[1]{{\color{blue}#1}}

\newcommand{\hide}[1]{}

\DeclareMathOperator*{\argmin}{argmin}
\DeclareMathOperator*{\argmax}{argmax}

\newcommand{\linkToPdf}[1]{\href{#1}{\blue{(pdf)}}}
\newcommand{\linkToPpt}[1]{\href{#1}{\blue{(ppt)}}}
\newcommand{\linkToCode}[1]{\href{#1}{\blue{(code)}}}
\newcommand{\linkToWeb}[1]{\href{#1}{\blue{(web)}}}
\newcommand{\linkToVideo}[1]{\href{#1}{\blue{(video)}}}
\newcommand{\linkToMedia}[1]{\href{#1}{\blue{(media)}}}
\newcommand{\award}[1]{\xspace} 

\newcommand{\myparagraph}[1]{{\bf#1.}}

\newcommand{\Euroc}{EuRoC\xspace}

\newcommand{\UncertaintyBound}{$0.1$\xspace}

\title{Probabilistic Volumetric Fusion \\ for Dense Monocular SLAM}

\author{
  Antoni Rosinol
  \and
  John J.~Leonard\\
  Massachusetts Institute of Technology \\
  {\tt\small \{arosinol, jleonard, lcarlone\}@mit.edu}
  \and
  Luca Carlone
}

\begin{document}

\maketitle
\begin{tikzpicture}[overlay, remember picture]
\path (current page.north east) ++(-6.3,-0.4) node[below left] {
Accepted for publication at WACV 2023, please cite as follows:
};
\end{tikzpicture}
\begin{tikzpicture}[overlay, remember picture]
\path (current page.north east) ++(-8.6,-0.8) node[below left] {
A. Rosinol, J. Leonard, L. Carlone
};
\end{tikzpicture}
\begin{tikzpicture}[overlay, remember picture]
\path (current page.north east) ++(-6.3,-1.2) node[below left] {
``Probabilistic Volumetric Fusion for Dense Monocular SLAM'',
};
\end{tikzpicture}
\begin{tikzpicture}[overlay, remember picture]
\path (current page.north east) ++(-5.2,-1.6) node[below left] {
 IEEE/CVF Winter Conf. on Applications of Computer Vision (WACV), 2023.
};
\end{tikzpicture}
\thispagestyle{empty}

\begin{abstract}
    We present a novel method to reconstruct 3D scenes from images by leveraging deep dense monocular SLAM and fast uncertainty propagation.
    The proposed approach is able to 3D reconstruct scenes densely,
    accurately, and in real-time while being robust to extremely noisy depth estimates coming from dense monocular SLAM.
    Differently from previous approaches,
    that either use ad-hoc depth filters, or that estimate the depth uncertainty from RGB-D cameras' sensor models,
    our probabilistic depth uncertainty derives directly from the information matrix of the underlying bundle adjustment problem in SLAM.
    We show that the resulting depth uncertainty provides an excellent signal to weight the depth-maps for volumetric fusion.
    Without our depth uncertainty, the resulting mesh is noisy and with artifacts,
    while our approach generates an accurate 3D mesh with significantly fewer artifacts.
    We provide results on the challenging Euroc dataset, and show that our approach achieves $92\%$ better accuracy
    than directly fusing depths from monocular SLAM, and up to $90\%$ improvements compared to the best competing approach.
\end{abstract}

\section{Introduction}
\label{sec:introduction}

\begin{figure*}[htbp]
    \centering
    \includegraphics[width=1.0\textwidth]{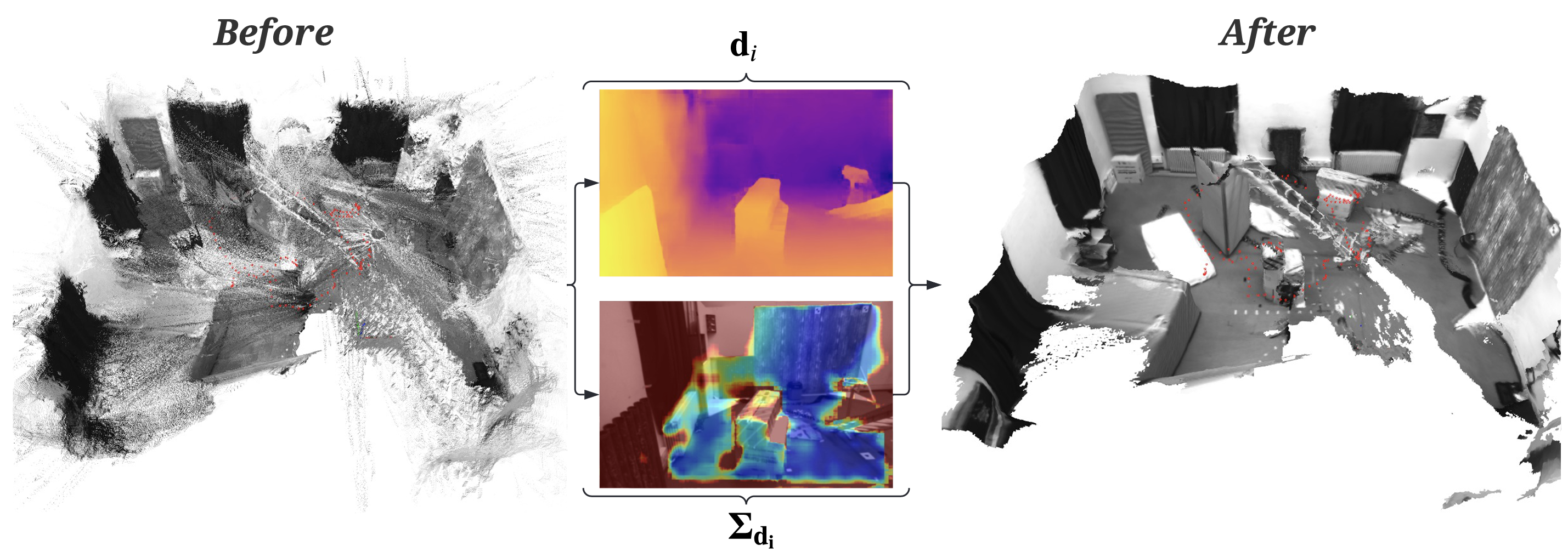}
    \caption{
      (Left) Raw 3D point-cloud generated from dense monocular SLAM by back-projecting the inverse depth-maps, without filtering or post-processing.
      (Right) Estimated 3D mesh after uncertainty-aware volumetric fusion of the depth-maps.
      Despite large amounts of noise in the depth-maps, the reconstructed 3D mesh using our proposed approach is accurate and complete.
      \Euroc  V2\_01 dataset.}
    \label{fig:main_fig}
\end{figure*}

3D reconstruction from monocular imagery remains one of the most difficult computer vision problems.
Achieving 3D reconstructions in real-time from images alone would enable many applications in robotics,
surveying, and gaming, such as autonomous vehicles, crop monitoring, and augmented reality.

While many 3D reconstruction solutions are based on RGB-D or Lidar sensors,
scene reconstruction from monocular imagery provides a more convenient solution.
RGB-D cameras can fail under certain conditions, such as under sunlight, and Lidar remains heavier and more expensive than a monocular RGB camera.
Alternatively, stereo cameras simplify the depth estimation problem to a 1D disparity search, but rely on accurate calibration of the cameras that 
is prone to miscalibration during practical operations.
Instead, monocular cameras are inexpensive, lightweight,
and represent the simplest sensor configuration to calibrate.

Unfortunately, monocular 3D reconstruction is a challenging problem due to the lack of explicit measurements of the geometry of the scene.
Nonetheless, great progress has been recently made towards monocular-based 3D reconstructions by leveraging deep-learning approaches.
Given that deep-learning currently achieves the best performance for optical flow \cite{teed2020raft}, and depth \cite{yao2018mvsnet} estimation,
 a plethora of works have tried to use deep-learning modules for SLAM.
For example, using depth estimation networks from monocular images \cite{Tateno17cvpr-CNN-SLAM}, multiple images, as in multi-view stereo \cite{koestler2022tandem},
or using end-to-end neural networks \cite{bloesch2018codeslam}.
However, even with the improvements due to deep-learning,
the resulting reconstructions are prone to errors and artifacts,
since the depth-maps are most of the time noisy and with outliers.

In this work, we show how we can 
substantially reduce the artifacts and inaccuracies in 3D reconstructions
from noisy depth maps estimated when using dense monocular SLAM.
To achieve this, we fuse the depth maps 
volumetrically by weighting each depth measurement
by its uncertainty, which we estimate probabilistically.
Differently from previous approaches,
we show that using the depth uncertainty derived from the information matrix of the bundle adjustment problem in monocular SLAM
leads to surprisingly accurate 3D mesh reconstructions.
Our approach achieves up to $90\%$ improvements in mapping accuracy, while retaining most of the scene geometry.

\myparagraph{Contributions}
    We show an approach to volumetrically fuse dense depth maps weighted by
     the uncertainties derived from the information matrix in dense SLAM. 
    Our approach enables the reconstruction of the scene up to a given maximum level of tolerable uncertainty.
    We can reconstruct the scene with superior accuracy compared to competing approaches,
        while running in real-time, and only using monocular images. %
        We achieve state-of-the-art 3D reconstruction performance in the challenging \Euroc dataset.

\section{Related Work}
\label{sec:related_work}

We review the literature on two different lines of work: dense SLAM and depth fusion.

\subsection{Dense SLAM}

The main challenges to achieve dense SLAM are 
(i) the computational complexity due to the shear amount of depth variables to be estimated, 
and (ii) dealing with ambiguous or missing information to estimate the depth of the scene, such as textureless surfaces or aliased images.

Historically, the first problem has been bypassed by decoupling the pose and depth estimation.
For example, DTAM \cite{Newcombe2011iccv-dtam} achieves dense SLAM by using the same paradigm as the sparse PTAM\cite{Klein07ismar},
 which tracked the camera pose first and then the depth, in a de-coupled fashion. %
The second problem is also typically avoided by using RGB-D or Lidar sensors, that provide explicit depth measurements, or stereo cameras that simplify depth estimation.

Nevertheless, recent research on dense SLAM has achieved impressive results in these two fronts.
To reduce the number of depth variables,
 CodeSLAM\cite{bloesch2018codeslam} 
optimizes instead the latent variables of an auto-encoder that infers 
depth maps from images. 
By optimizing these latent variables, the dimensionality of the problem is significantly reduced, while the resulting depth maps remain dense.
Tandem\cite{koestler2022tandem} is able to reconstruct 3D scenes 
with only monocular images by using a pre-trained MVSNet-style 
neural-network on monocular depth estimation,
and then decoupling the pose/depth problem by performing frame-to-model photometric tracking.
Droid-SLAM\cite{teed2021droid} shows that by adapting a state-of-the-art dense optical flow estimation architecture \cite{teed2020raft} to the 
problem of visual odometry, it is possible to achieve competitive results in a variety of challenging datasets 
(such as the Euroc \cite{Burri16ijrr-eurocDataset} and TartanAir \cite{wang2020tartanair} datasets),
even though it requires global bundle-adjustment to outperform model-based approaches.
Droid-SLAM avoids the dimensionality problem by using downsampled depth maps
that are subsequently upsampled using a learned upsampling operator.
Finally, there are a myriad of works that avoid the dimensionality and ambiguity problems stated above that have nevertheless recently achieved improved performance.
For example, iMap\cite{sucar2021imap} and Nice-SLAM\cite{zhu2022nice} 
can build accurate 3D reconstructions by both de-coupling pose and depth estimates and using RGB-D images, and achieve photometrically accurate reconstructions by using neural radiance fields \cite{mildenhall2020nerf}.
Given these works, we can expect future learned dense SLAM to become more accurate and robust.

Unfortunately, we are not yet able to achieve pixel-perfect depth maps from casual image collections, and fusing these depth maps directly into a volumetric representation often leads to artifacts and inaccuracies.
Our work leverages Droid-SLAM\cite{teed2021droid} to estimate extremely dense 
(but very noisy) depth maps per keyframe (see left pointcloud in \cref{fig:main_fig}), 
that we successfully fuse into a volumetric representation by weighting the depths by their uncertainty, estimated as marginal covariances.

\subsection{Depth Fusion}

The vast majority of 3D reconstruction algorithms are based on fusing depth maps provided from a depth-sensor 
into a volumetric map \cite{Newcombe2011ismar-kinectfusion, Oleynikova17iros-voxblox,Rosinol20icra-Kimera}.
Most of the literature using volumetric representations have therefore focused on studying ways to 
obtain better depth maps, such as with post-processing techniques,
or on the weighting function to be used when fusing the depths \cite{Newcombe2011ismar-kinectfusion,Whelan15rss-elasticfusion,bylow2013real,nguyen2012modeling}.
Most of the literature, by assuming that the depth maps come from a sensor, have focused on sensor modelling.
Alternatively, when using deep-learning, a similar approach is to make a neural network learn the weights instead.
For example, RoutedFusion\cite{weder2020routedfusion} and NeuralFusion\cite{weder2021neuralfusion} learn to de-noise volumetric reconstructions from RGB-D scans.

In our case, since the depth maps are estimated through dense bundle adjustment,
we propose to directly fuse the depth maps using the marginal covariances of the estimated depths.
This is computationally difficult to do since, in dense SLAM,
the number of depths per keyframe can be as high as the total number of pixels in the frame ($\approx 10^5$). 
We show below how we can achieve this by leveraging the block-sparse structure of the information matrix.

\section{Methodology}

The main idea of our method is to fuse extremely dense but noisy depth-maps
weighted by their probabilistic uncertainty into a volumetric map, and then extract a 3D mesh that has a given maximum uncertainty bound.
Towards this goal, we leverage Droid-SLAM's formulation to produce pose estimates and dense depth-maps,
and extend it to generate dense uncertainty-maps.

We will first show how we compute depth uncertainties from the information matrix of the underlying bundle adjustment problem efficiently.
Then, we present our fusion strategy to produce a probabilistically sound volumetric map.
Finally, we show how we extract a mesh from the volume at a given maximum uncertainty bound.

\subsection{Dense Monocular SLAM}

At its core, classical vision-based inverse-depth indirect SLAM 
solves a bundle adjustment (BA) problem 
where the 3D geometry is parametrized as a set of (inverse) depths per keyframe.
This parametrization of the structure leads to an extremely efficient way of solving the dense BA problem,
 which can be decomposed into the familiar arrow-like block-sparse matrix with cameras and depths in sequence:
\begin{equation}
    \label{eq:hessian}
    H\mathbf{x} = \mathbf{b}, \quad \text{\ie} \quad 
    \left[
        \begin{array}{cc}
            C & E \\
            E^{T} & P
        \end{array}
    \right]
    \left[
        \begin{array}{l}
            \Delta \boldsymbol{\xi} \\
            \Delta \mathbf{d}
        \end{array}
    \right]
    =
    \left[
        \begin{array}{c}
            \mathbf{v} \\
            \mathbf{w}
        \end{array}
    \right],
\end{equation}
where $H$ is the Hessian matrix, $C$ is the block camera matrix,
and $P$ is the diagonal matrix corresponding to the points (one inverse depth per pixel per keyframe).
We represent by $\Delta\boldsymbol{\xi}$ the delta updates on the lie algebra of the camera poses in $SE(3)$,
while $\Delta\mathbf{d}$ is the delta update to the per-pixel inverse depths.

To solve the BA problem, the Schur complement of the Hessian $H$ with respect to $P$ (denoted as $H / P$) is first calculated to eliminate the inverse depth variables:
\begin{equation}
    \label{eq:schur}
    (H / P) \Delta \boldsymbol{\xi} = \left[C-E P^{-1} E^{T}\right]  \Delta \boldsymbol{\xi}=\left(\mathbf{v}-E P^{-1} \mathbf{w}\right).
\end{equation}
The Schur complement can be quickly computed given that $P^{-1}$ consists on an element-wise inversion of each diagonal element that can be performed in parallel,
since $P$ is a large but diagonal matrix.

The resulting matrix $(H/P)$ is known as the reduced camera matrix.
The system of equations in \cref{eq:schur} only depends on the keyframe poses.
Hence, we first solve for the poses using the Cholesky decomposition of $(H/P) = LL^T$ using front and then back-substitution.
The resulting pose solutions $\Delta \boldsymbol{\xi}$ are then used to solve back for the inverse depth maps $\Delta \mathbf{d}$, as follows:
\begin{equation}
    \label{eq:solve_ba_depth}
    \begin{aligned}
        &\Delta \mathbf{d}=P^{-1}\left(\mathbf{w}-E^{T} \Delta \boldsymbol{\xi}\right).
    \end{aligned}
\end{equation}

Nevertheless, to make inference fast enough for real-time SLAM,
the inverse depth-maps are estimated at a lower $1/8^{\text{th}}$ resolution than the original images,
in our case of $69 \times 44$ pixels (Euroc dataset's original resolution is $752 \times 480$ which we first downsample to $512 \times 384$).
Once this low resolution depth map of is solved for, a learned upsampling operation (shown first in \cite{teed2020raft} for optical flow estimation, and used as well in Droid-SLAM)
recovers the full resolution depth map.
This allows us to effectively reconstruct dense depth maps of the same resolution as the input images.

Solving the same BA problem with high resolution depth maps is prohibitively expensive for real-time SLAM,
 the computation of depth-uncertainties further exacerbates the problem.
We believe this is the reason why other authors have not used depth uncertainties derived from BA for real-time volumetric 3D reconstruction:
using full-depth BA is prohibitively expensive, and using sparse-depth BA leads to way too sparse depth-maps for volumetric reconstruction.
The alternative has always been to use sparse BA for pose estimation and a first guess of the geometry, followed by a densification step unrelated to the information matrix in sparse BA \cite{Schonberger16cvpr-SfMRevisited}.
That is why other authors have proposed to use alternative 3D representations for dense SLAM,
such as latent vectors in CodeSLAM\cite{bloesch2018codeslam}.
Our approach can also be applied to CodeSLAM.

\subsection{Inverse Depth Uncertainty Estimation}

Given the sparsity pattern of the Hessian, we can extract the required marginal covariances for the per-pixel depth variables efficiently.
The marginal covariances of the inverse depth-maps $\mathbf{\Sigma}_{\mathbf{d}}$ are given by:
\begin{equation}
    \label{eq:cov}
    \begin{aligned}
        \mathbf{\Sigma}_{\mathbf{d}} &= P^{-1}+P^{-1} E^T\mathbf{\Sigma}_{\mathbf{T}} E P^{-1} \\
        \mathbf{\Sigma}_{\mathbf{T}} &= (H / P)^{-1}, \\
    \end{aligned}
\end{equation}
where $\mathbf{\Sigma}_{\mathbf{T}}$ is the marginal covariance of the poses.
Unfortunately, a full inversion of $H/P$ can be costly to compute.
Nevertheless, since we already solved the original BA problem by factorizing $H/P$ into its Cholesky factors, 
 we can re-use them in the following way, similarly to \cite{ila2017fast}:
\begin{equation}
    \begin{aligned}
        \mathbf{\Sigma}_{\mathbf{d}} &= P^{-1} + P^{-1} E^T\mathbf{\Sigma}_{\mathbf{T}} E P^{-1} \\ 
                                    &= P^{-1} + P^{-1} E^T(LL^T)^{-1} E P^{-1} \\ 
                                    &= P^{-1} + P^{-T} E^T L^{-T} L^{-1} E P^{-1} \\
                                    &= P^{-1} + F^TF,\\
    \end{aligned}
\end{equation}
where $F=L^{-1}EP^{-1}$.
Hence, we only need to invert the lower triangular Cholesky factor $L$, which is a fast operation to compute by substitution. 
Therefore, we can compute all inverse matrices efficiently: 
the inverse of $P$ is given by the element-wise inversion of each diagonal entry,
and we avoid a full inversion of $(H/P)$ by inverting instead its Cholesky factor.
It then suffices to multiply and add matrices together:
\begin{equation}
    \begin{aligned}
        [\mathbf{\Sigma_{d}}]_{i} = \sigma^2_{d_i}
        = P^{-1}_{i} + \{F^T F\}_{i} 
        = P^{-1}_{i} + \sum_k F_{ki}^2,
    \end{aligned}
\end{equation}
where $d_{i}$ is one of the per-pixel inverse depths.
Since most of the operations can be computed in parallel, we leverage the massive parallelism of GPUs.

\subsection{Depth Upsampling \& Uncertainty Propagation}
Finally, since we want to have a depth-map of the same resolution than the original images,
 we upsample the low-resolution depth-maps using the convex upsampling operator defined in Raft\cite{teed2020raft} and also used in Droid\cite{teed2021droid}.
This upsampling operation calculates a depth estimate for each pixel in the high-resolution depth-map by taking the convex combination of the neighboring depth values in the low-resolution depth-map.
The resulting depth estimates are given for each pixel by:
\begin{equation}
    \label{eq:upsampling}
    d = \sum_{i=0}^8 w_i d_i,
\end{equation}
where the $w_i$ are learned weights (more details can be found in Raft\cite{teed2020raft}),
 and $d_i$ is the inverse depth of a pixel in the low-resolution inverse depth-map surrounding the pixel for which we are computing the depth (a $3\times 3$ window is used to sample neighboring depth values).

Assuming independence between inverse depth estimates, the resulting inverse depth variance is given by:
\begin{equation}
   \sigma_d^2 = \sum_{i=0}^8 w_i^2 \sigma^2_{d_i}, 
\end{equation}
where $w_i$ are the same weights used for the inverse depth upsampling in \cref{eq:upsampling},
 and $\sigma^2_{d_i}$ is the variance of the inverse depth of a pixel in the lower resolution inverse depth-map surrounding the pixel to be calculated.
We upsample the inverse depths and uncertainties by a factor of $8$, going from a $69\times 44$ resolution to a $512 \times 384$ resolution.

So far we have been working with inverse depths, the last step is to convert them to actual depth and depth-variances.
We can easily compute the depth variance by using nonlinear uncertainty propagation: 
\begin{equation}
    \label{eq:sigma_z}
    z = 1 / d, \quad \quad \sigma_z = \sigma_d / d^2,
\end{equation}
where $z$ is the resulting depth, and $d$ is the inverse depth.

\begin{figure*}[t!]
    \centering
    \includegraphics[width=1.0\textwidth]{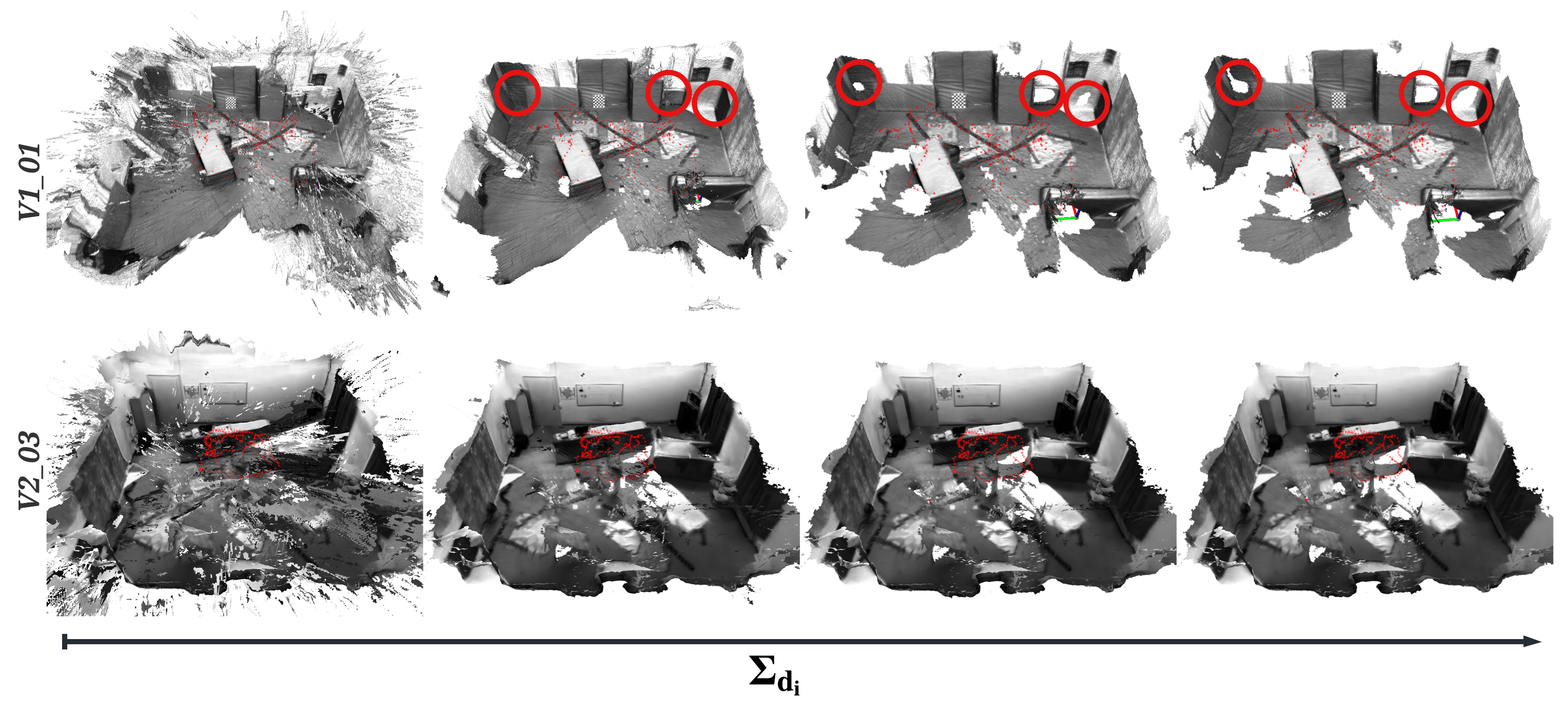}
    \caption{3D mesh reconstructions for a given maximum admissible mesh uncertainty $\mathbf{\Sigma_{d_i}}$
     logarithmically decreasing from an infinity upper-bound (\ie minimum weight of $0.0$, left-most 3D mesh) to $0.01$ (\ie minimum weight of $10$, right-most 3D mesh).
     The regions highlighted with red circles disappear first because of high uncertainty. These correspond to textureless and aliased regions.
     The two closest red circles correspond to the same region as the one depicted in \cref{fig:aliasing}.}
    \label{fig:threshold}
\end{figure*}

\subsection{Uncertainty-aware Volumetric Mapping}

Given the dense depth-maps available for each keyframe, it is possible to build a dense 3D mesh of the scene.
Unfortunately, the depth-maps are extremely noisy due to their density, since even textureless regions are given a depth value.
Volumetrically fusing these depth-maps reduces the noise,
but the reconstruction remains inaccurate and corrupted by artifacts
(see `Baseline' in \cref{fig:sota}, which was computed by fusing the pointcloud shown in \cref{fig:main_fig}).

While it is possible to manually set filters on the depth-maps 
(see PCL's documentation for examples of possible depth filters\cite{Rasu11icra}) and 
Droid implements one ad-hoc depth filter (see Droid in \cref{fig:sota}),
we propose to use instead the estimated depth maps' uncertainties,
which provide a robust and mathematically sound way to reconstruct the scene.

Volumetric fusion is grounded on a probabilistic model \cite{Curless96siggraph},
whereby each depth measurement is assumed to be independent and Gaussian distributed.
Under this formulation, the signed distance function (SDF) $\phi$, which we try to estimate, maximizes the following likelihood:
\begin{equation}
    \phi^\star = \argmax_\phi \quad p\left(z, \sigma_z \mid \phi\right),
\end{equation}
\begin{equation}
    p\left(z, \sigma_z \mid \phi\right) \propto \prod_{i} \exp \left(-\frac{1}{2} \|\phi-z_{i}\|^{2}_{\sigma^2_{z_i}}\right).
\end{equation}
Taking the negative log leads to a weighted least-squares problem:
\begin{equation}
    \phi^\star = \argmin_\phi
    \frac{1}{2} \sum_{i} \frac{\left(\phi-z_{i}\right)^{2}}{\sigma_{z_i}},
\end{equation}
the solution of which is obtained by setting the gradient to zero and solving for $\phi$,
 leading to a weighted average over all the depth measurements:
\begin{equation}
    \label{eq:weights}
    \begin{aligned}
    \phi&=\frac{\sum_{i} z_{i} / \sigma_{z_i}}
               {\sum_{i} 1 / \sigma_{z_i}}
         =\frac{\sum_{i} w_i z_{i}}{\sum_{i} w_i},
    \end{aligned}
\end{equation}
with the weights $w_i$ defined as $w_i = \sigma_{z_i}^{-1}$.

In practice, the weighted average is computed incrementally for every new depth-map,
by updating the voxels in the volume with a running average,
leading to the familiar volumetric reconstruction equations:
\begin{equation}
    \phi_{i+1} =\frac{W_{i} \phi_{i}+w_i z_i}{W_{i}+w_i}, \quad  W_{i+1} = W_{i}+w_i,
\end{equation}
where $W_{i}$ is the weight stored in each voxel.
The weights are initialized to zero, $W_{0} = 0$,
 and the TSDF is initialized to the truncation distance $\tau$,
 $\phi_0 = \tau$ (in our experiments $\tau = 0.1$m). 
The formulation above, as a running weighted average, 
is extremely flexible in terms of the weight function to be used.
This flexibility has led to many different ways to fuse depth-maps,
sometimes departing from its probabilistic formulations.

Most approaches determine a weight function by 
modelling the error distribution from the depth-sensor used,
be it a laser scanner, an RGB-D camera, or a stereo camera \cite{Curless96siggraph,Oleynikova17iros-voxblox,Rosinol21ijrr-Kimera}.
For example, Nguyen et al. \cite{nguyen2012modeling} modeled the residual errors from an RGB-D camera 
and determined that the depth variance was dominated by $z^{2}$, $z$ being the measured depth.
Bylow et al. \cite{bylow2013real} analyzed a variety of weight functions,
and concluded that a linearly decreasing weight function behind the surface led to the best results.
Voxblox \cite{Oleynikova17iros-voxblox} combined these two works into a simplified formulation with good results, 
which is also used in Kimera \cite{Rosinol20icra-Kimera}.

In our case, there is no sophisticated weighting or sensor model needed;
the depth uncertainties are computed from the inherently probabilistic factor-graph formulation in SLAM.
Specifically, our weights are inversely proportional to the marginal covariance of the depths, as derived 
from a probabilistic perspective in \cref{eq:weights}.
Notably, these weights result from the fusion of hundreds of optical flow 
measurements with their associated measurement noise estimated by a neural network (GRU's output in Droid\cite{teed2021droid}).

\subsection{Meshing with Uncertainty Bounds}

Given that our voxels have a probabilistically sound uncertainty estimate
of the signed distance function, we can extract the iso-surface for different levels of maximum uncertainty allowed.
We extract the surfaces using marching cubes, by only meshing those voxels which have an uncertainty estimate below the maximum allowed uncertainty.
The resulting mesh has only geometry with a given upper-bound uncertainty,
while our volume contains all the depth-maps' information.

If we set the uncertainty bound to infinity \ie, a weight of $0$,
 we recover the baseline solution, which is extremely noisy.
By incrementally decreasing the bound, we can balance between having more accurate, but less complete 3D meshes, and vice-versa.
In \Cref{sec:results}, we show different meshes obtained with decreasing values for the uncertainty bound (\cref{fig:threshold}).
In our experiments, we did not try to find a particular pareto optimal solution for our approach, but instead used a fixed maximum upper bound on the uncertainty of \UncertaintyBound,
which leads to very accurate 3D meshes with a minor loss in completeness (see \cref{sec:results} for a quantitative evaluation).
Note that, without fixing the scale, this uncertainty bound is unitless,
 and may need to be adapted depending on the estimated scale.

\subsection{Implementation Details}

We perform all computations in Pytorch with CUDA,
and use an RTX 2080 Ti GPU for all our experiments (11Gb of memory).
For the volumetric fusion, we use Open3D's \cite{Zhou18arxiv-open3D} library,
that allows for custom volumetric integration.
We use the same GPU for SLAM and to perform volumetric reconstruction.
We use the pre-trained weights from Droid-SLAM \cite{teed2021droid}.
Finally, we use the marching cubes algorithm implemented in Open3D to extract the 3D mesh.

\section{Results}
\label{sec:results}

\Cref{ssec:qual_mapping_quality} and \cref{ssec:quan_mapping_quality} 
show a qualitative and quantitative evaluation of our proposed 3D mesh reconstruction algorithm,
with respect to the baseline and state-of-the-art approaches,
on the \Euroc dataset,
using the subset of scenes that have a ground-truth pointcloud.

The qualitative analysis presents the strengths and weaknesses of our approach, 
and compares with other techniques in terms of perceptual quality and geometric fidelity,
For the quantitative part,
we compute the RMSE for both accuracy and completeness metrics,
to objectively assess the performance of our algorithm against competing approaches. 
We now describe the dataset and different approaches used for evaluation.

\subsection{Datasets \& Methods for Evaluation}
\label{ssec:datasets}

For evaluation of our reconstruction algorithm, we use the \Euroc dataset, which consists of images recorded from a drone flying in an indoor space.
We use the ground-truth pointclouds available in the \Euroc \texttt{V1} and \texttt{V2} datasets to assess the quality of the 3D meshes
produced by our approach. For all our experiments, we set our maximum admissible mesh uncertainty to \UncertaintyBound.

We compare our approach with two different open-source state-of-the-art learning and model-based dense VO algorithms:
Tandem\cite{koestler2022tandem}, a learned dense monocular VO algorithm that uses a MVSNet-style architecture and photometric bundle-adjustment, and
Kimera\cite{Rosinol20icra-Kimera}, a model-based dense stereo VIO algorithm.
Both use volumetric fusion to reconstruct the 3D scene and output a 3D mesh of the environment.
We also present the results from fusing Droid's pointclouds after Droid's ad-hoc depth filter,
which computes the support of a depth value by counting the number of nearby depth-maps that reproject within a threshold ($0.005$ by default).
Any depth value with less than $2$ supporting depths, or smaller than half of the mean depth, is then discarded.
Droid's filter is used to remove outliers on the depth-maps, while we fuse all depth-maps weighted by their uncertainty.
As our baseline, we use the raw pointclouds estimated by Droid,
and fuse them directly into a volumetric reconstruction.

\begin{figure}[htbp]
    \centering
    \includegraphics[width=1.0\columnwidth]{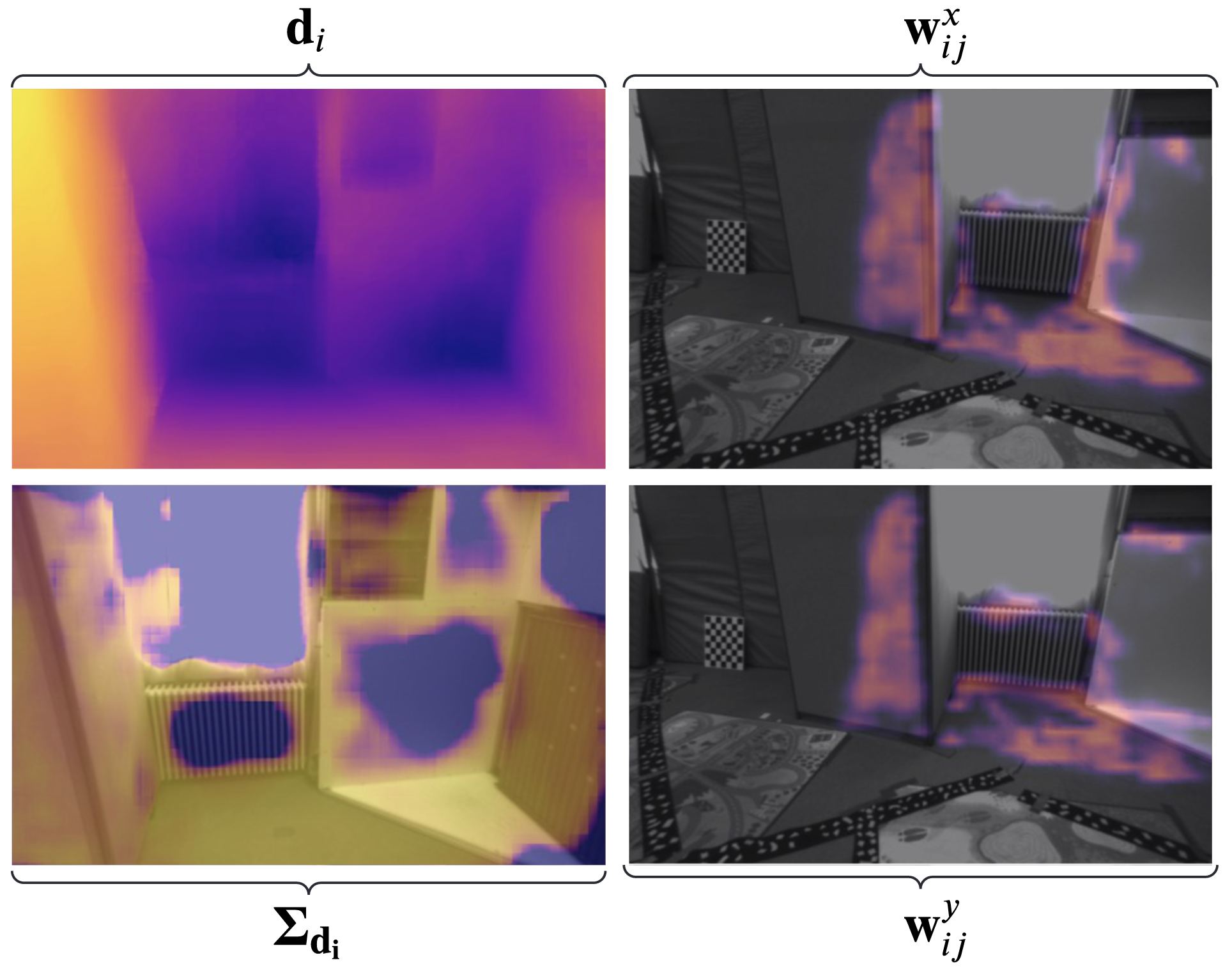}
    \caption{
      (Left Column) Frame $i$.
      (Right Column) Frame $j$.
      (Top-Left) Estimated depth-map for frame $i$.
      (Bottom-Left) Estimated depth-map uncertainty for frame $i$.
      (Top-Right) Optical-flow measurement weights for the $x$ component of the flow from frame $i$ to frame $j$.
      (Bottom-Right) Optical-flow measurement weights for the $y$ component.
      Note that the flow weights are localized where frame $i$ is visible in frame $j$.
      The depth uncertainty results from the fusion of several optical-flow measurements, rather than a single one.
      For the left column, low values are in yellow, high values are in blue.
      For the right column, low values are in blue, high values are in yellow.
      \Euroc  V1\_01 dataset.}
    \label{fig:aliasing}
\end{figure}

\subsection{Qualitative Mapping Performance}
\label{ssec:qual_mapping_quality}

\begin{figure*}[t!]
    \centering
    \includegraphics[trim={0cm 0cm 0cm 0cm}, clip, width=1.0\textwidth]{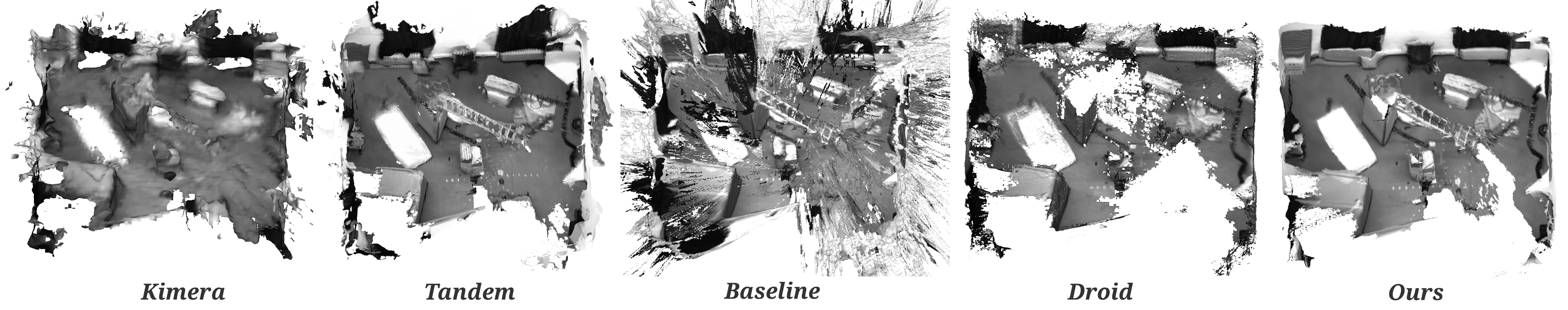}
    \caption{Comparison of the 3D mesh reconstructed by Kimera\cite{Rosinol20icra-Kimera}, Tandem\cite{koestler2022tandem}, our baseline, and Droid's depth filter \cite{teed2021droid} (using the default threshold of $0.005$) versus our approach using a maximum tolerated mesh uncertainty of \UncertaintyBound. \Euroc  V2\_01 dataset.}
    \label{fig:sota}
\end{figure*}

\Cref{fig:threshold} shows how we can trade-off accuracy for completeness by varying the maximum level of uncertainty allowed in the 3D reconstruction.
We can also see how the less certain geometry gradually disappears.
The least certain geometry corresponds to the artifacts floating in 3D space due to the depths that are poorly triangulated, and scattered in 3D rays when back-projected (first column in \cref{fig:threshold}).
Then, we see that the subsequent geometry that disappears corresponds to textureless regions (left-most and right-most red circles in each column \cref{fig:threshold}).
Interestingly, the removed geometry that follows after textureless regions corresponds to highly aliased regions (middle red circles in each column \cref{fig:threshold}),
such as the heaters, or the center of the checkerboards present in the room.

A careful look in \cref{fig:aliasing} shows that the estimated depth uncertainty $\mathbf{\Sigma_{d}}$ is not only large for textureless regions,
but also for regions with strong aliasing that are difficult to resolve for optical-flow based SLAM algorithms (the heater in the middle of the image).
Indeed, the optical flow weights (right column in \cref{fig:aliasing}) are close to $0$ for regions with strong aliasing or textureless regions.
This emerging behavior is an interesting result that could be used to detect aliased geometry, or to guide hole-filling reconstruction approaches.

\begin{table}[t!]
    \label{tab:accuracy}
    \centering
    \caption{\textbf{Accuracy RMSE [m]}: for the 3D mesh generated from 
             our approach compared to Kimera, Tandem, Droid's filter, and our baseline,
             on the subset of the \Euroc dataset with ground-truth pointclouds.
             Note that if an approach only estimates a few accurate points (\eg Droid), the accuracy can reach $0$.
             Best approach in bold, second-best in italics, $-$ indicates no mesh reconstructed.
             }
    \vspace{1em}
    \begin{tabular}{ccccccc}
      \toprule
       & \multicolumn{3}{c}{V1} & \multicolumn{3}{c}{V2} \\
      \cmidrule(l{1pt}r{1pt}){2-4} \cmidrule(l{1pt}r{1pt}){5-7}
        & 01 & 02 & 03 & 01 & 02 & 03 \\
      \midrule
      \makecell{Kimera}        & 0.14          & 0.15          & \textit{0.16} & 0.22          & 0.22          & 0.25          \\
      \makecell{Tandem}        & 0.10          & \textit{0.12} & 0.20          & 0.12          & \textit{0.16} & 0.29          \\
      \makecell{Baseline}      & 0.22          & 0.24          & 0.28          & 0.32          & 0.31          & 0.34          \\
      \makecell{Droid}         & \textit{0.05} & $-$           & $-$           & \textit{0.07} & $-$           & \textit{0.11} \\
      \makecell{\textbf{Ours}} & \textbf{0.03} & \textbf{0.03} & \textbf{0.02} & \textbf{0.04} & \textbf{0.04} & \textbf{0.07} \\
      \bottomrule
    \end{tabular}
\end{table}{}

\Cref{fig:sota} qualitatively compares the 3D reconstruction of Kimera\cite{Rosinol20icra-Kimera}, Tandem\cite{koestler2022tandem}, the baseline approach, Droid\cite{teed2021droid} and our approach.
We can see that compared to our baseline approach, we perform much better both in terms of accuracy and completeness.
Kimera is able to build a complete 3D reconstruction but lacks both accuracy and detail compared to our approach.
Tandem is the competing approach performing best, and results in similar reconstructions than our proposed approach.
From \cref{fig:sota}, we can see that Tandem is more complete than ours (see bottom right missing strip of floor in our reconstruction),
while being slightly less accurate (see top-left section of the reconstruction that is distorted in Tandem's mesh).
In principle, our approach could reconstruct the ground-floor of the room as well (the baseline reconstruction has that information).
Nonetheless, in a robotics context, it is better to be aware of what region is unknown rather than committing with a first guess that is inaccurate,
since that can close path-ways that could have been traversed by the robot (a common scenario in the DARPA SubT challenge \cite{agha2021nebula}, where robots explore a network of tunnels and caves).
Finally, Droid's depth filter is missing important regions and negatively affects the reconstruction accuracy.

\subsection{Quantitative Mapping Performance}
\label{ssec:quan_mapping_quality}

\begin{table}[t!]
    \label{tab:completeness}
    \centering
    \caption{\textbf{Completeness RMSE [m]}: for the 3D mesh generated from 
             our approach compared to Kimera, Tandem, and our baseline,
             on the subset of the \Euroc dataset with ground-truth pointclouds.
             Note that if an approach estimates a dense cloud of points (\eg Baseline), completeness can reach $0$.
             Best approach in bold, second-best in italics, $-$ indicates no mesh reconstructed.
             }
    \vspace{1em}
    \begin{tabular}{ccccccc}
      \toprule
       & \multicolumn{3}{c}{V1} & \multicolumn{3}{c}{V2} \\
      \cmidrule(l{1pt}r{1pt}){2-4} \cmidrule(l{1pt}r{1pt}){5-7}
        & 01 & 02 & 03 & 01 & 02 & 03 \\
      \midrule
      \makecell{Kimera}        & 0.36          & 0.38          & 0.35          & 0.48          & 0.43          & 0.41          \\
      \makecell{Tandem}        & \textit{0.16} & \textit{0.12} & \textit{0.13} & \textit{0.20} & \textit{0.12} & 0.24          \\
      \makecell{Baseline}      & \textbf{0.05} & \textbf{0.04} & \textbf{0.04} & \textbf{0.05} & \textbf{0.05} & \textbf{0.05} \\
      \makecell{Droid}         & 0.62          & $-$           & $-$           & 0.35          & $-$           & 0.32           \\
      \makecell{\textbf{Ours}} & 0.20          & 0.16          & 0.19          & 0.24          & 0.18          & \textit{0.16}  \\
      \bottomrule
    \end{tabular}
\end{table}{}

We evaluate each mesh against the ground truth using the 
\textit{accuracy} and \textit{completeness} metrics, 
as in~\cite[Sec. 4.3]{Rosinol18thesis}:
(i) we first compute a point cloud by sampling 
the reconstructed 3D mesh with a uniform density of $10^4~\text{points}/\text{m}^2$, 
(ii) we register the estimated 
and the ground truth clouds with ICP~\cite{Besl92pami} using \emph{CloudCompare}~\cite{CloudCompare}, 
and 
(iii) we evaluate the average distance from ground truth point cloud to its nearest neighbor in the estimated point cloud (accuracy), and vice-versa (completeness), with a $0.5$m maximum distance.

\Cref{tab:accuracy} and \cref{tab:completeness} provide a quantitative comparison 
between our proposed approach, Droid's filter, and our baseline,
as well as a comparison with Kimera\cite{Rosinol20icra-Kimera} and Tandem \cite{koestler2022tandem}, in terms of accuracy and completeness.
As we can see from the tables, our proposed approach is the best performing in terms of accuracy by a substantial margin (as high as $90\%$ compared to Tandem 
and $92\%$ compared to the baseline for V1\_03), while Tandem achieves the second-best accuracy overall.
In terms of completeness, Tandem achieves the best performance (after the baseline approach), followed by our approach.
Droid's filter achieves good accuracy at the expense of substantially incomplete meshes.

\begin{figure}[htbp]
    \centering
    \includegraphics[width=0.8\columnwidth]{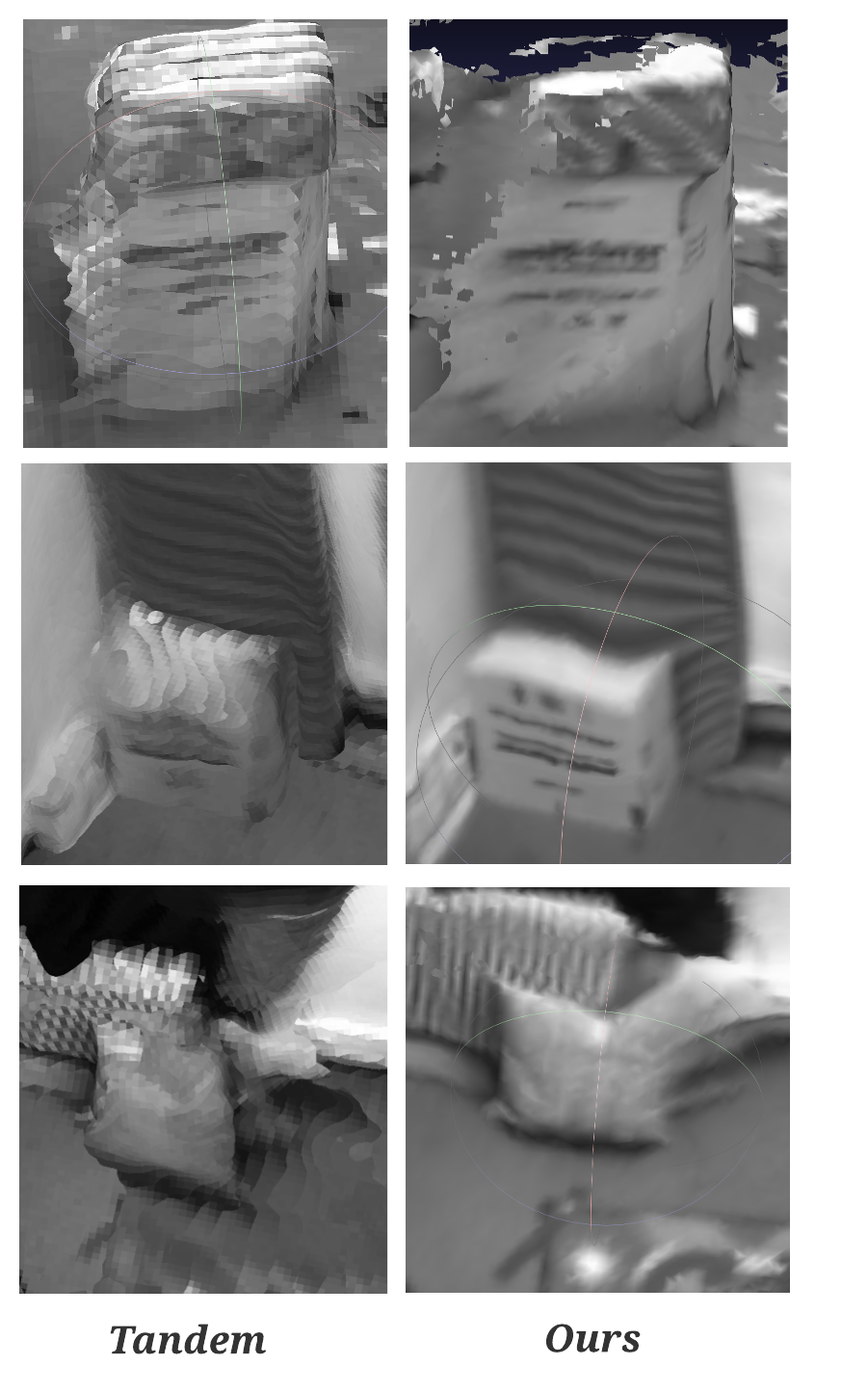}
    \caption{Closer look at the differences between Tandem's 3D reconstructions and ours.
    \Euroc  V2\_01 dataset.}
    \label{fig:close_look}
\end{figure}

\begin{figure}[htbp]
    \centering
    \includegraphics[width=0.98\columnwidth]{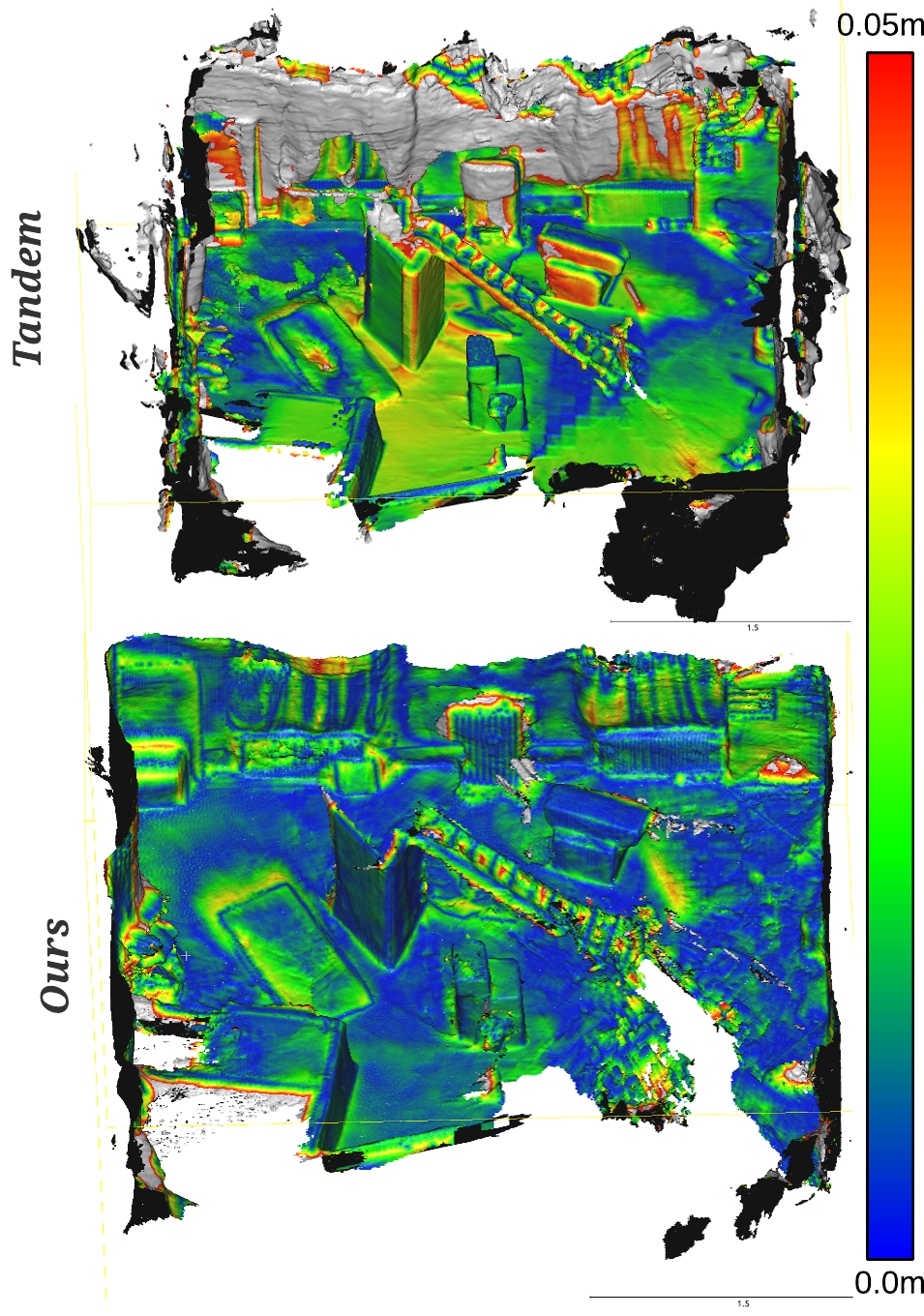}
    \caption{ Accuracy evaluation of both Tandem's (top) and our (bottom) 3D mesh reconstruction results.
    We truncate the color scale at $0.05$m, and visualize anything above it as gray, up to $0.5$m (geometry beyond this error is discarded).
    Note that our reconstruction has its biggest error at $0.37$m, while Tandem's largest error is beyond the $0.5$m bound.
     \Euroc  V2\_01 dataset.}
    \label{fig:accuracy_mesh}
\end{figure}

\Cref{fig:accuracy_mesh} shows the estimated cloud (V2\_01) 
color-coded by the distance to the closest point
in the ground-truth cloud (accuracy) for both Tandem (top) and our reconstruction (bottom).
We can see from this figure where our reconstruction is more accurate than Tandem's.
In particular, 
it is interesting to see that Tandem tends to generate inflated geometry,
 particularly in textureless regions such as the black curtains in the V2\_01 dataset (gray-colored geometry).
Our approach has better details, and better overall accuracy.
\Cref{fig:close_look} shows a close-up view of the reconstructed 3D mesh for both Tandem and our approach.
Our reconstructions tend to be less complete and suffer from bleeding edges,
but retain most of the details, while Tandem's reconstructions lack overall detail and tend to be slightly inflated,
but remain more complete.

\subsection{Real-Time Performance}

Downsampling the Euroc images to $512 \times 384$ resolution leads to tracking speeds of $15$ frames per second.
Computing the depth uncertainties decreases the tracking speed by a few frames per second to $13$ frames per second.
Volumetrically fusing the depth estimates, with or without depth uncertainties, takes less than $20$ms.
Overall, our pipeline is able to reconstruct the scene in real-time at $13$ frames per second,
 by parallelizing camera tracking and volumetric reconstruction, and by using custom CUDA kernels.

\section{Conclusion}
\label{sec:conclusions}

We propose an approach to 3D reconstruct scenes using dense monocular SLAM
and fast depth uncertainty computation and propagation.
We show that our depth-map uncertainties are a
source of reliable information for accurate and complete
3D volumetric reconstructions, resulting in meshes that have significantly lower noise and artifacts.

Given the mapping accuracy and probabilistic uncertainty estimates afforded by our approach,
we can foresee future research to focus on active exploration of uncertain regions in the map,
reconstructing the 3D scene beyond its geometry by
incorporating semantics, as in Kimera-Semantics \cite{Rosinol21ijrr-Kimera},
or by using neural volumetric implicit representations
for photometrically-accurate 3D reconstructions, as in Nice-SLAM\cite{zhu2022nice}.

\section*{Acknowledgments}
  This work is partially funded by `la Caixa' Foundation (ID 100010434), LCF/BQ/AA18/11680088 (A. Rosinol), `Rafael Del Pino' Foundation (A. Rosinol), 
  ARL DCIST CRA W911NF-17-2-0181, and ONR MURI grant N00014-19-1-2571.
  We thank Bernardo Aceituno for helpful discussions.

\addtolength{\textheight}{-2cm}  %

{\small
\bibliographystyle{ieee_fullname}
\bibliography{references/refs,references/myRefs}
}

\end{document}